\documentclass[11pt,a4paper]{article}
\usepackage[hyperref]{naaclhlt2019}
\usepackage{times}
\usepackage{latexsym}

\usepackage{url}

\usepackage{amsfonts}
\usepackage{amsmath}
\usepackage{amssymb}
\usepackage{graphicx}
\usepackage{stmaryrd}
\usepackage[shortlabels]{enumitem}
\usepackage{lipsum}
\usepackage{booktabs}

\aclfinalcopy 

\usepackage{multirow}
\usepackage{marvosym}
\usepackage{xspace}

\newcommand\ie{\emph{i.e.}}

\newcommand\discofuse{\textsc{DiscoFuse}}
\newcommand\dfs{\textsc{DfSport}}
\newcommand\dfw{\textsc{DfWiki}}
\newcommand\dfsw{\textsc{DfS+W}}
\newcommand\websplit{\textsc{WebSplit}\xspace}
\newcommand\exact{\emph{Exact}}
\newcommand\sari{\emph{SARI}}
\newcommand\jb[1]{\textcolor{red}{[JB: #1]}}
\newcommand\mg[1]{\textcolor{blue}{MG: #1}}

\newcommand\nl[1]{{\it``#1''}}

\newcommand\ignore[1]{}
\newcommand{\angleb}[1]{$\langle \text{#1} \rangle$}

\title{\textsc{DiscoFuse}: A Large-Scale Dataset for Discourse-Based \\ Sentence Fusion}

\author{Mor Geva \thanks{\hspace{5px}Work done during internship at Google AI.} \\
Tel Aviv University\\
\tt\small{morgeva@mail.tau.ac.il}\\\And
  Eric Malmi \\
Google AI\\
\tt\small{emalmi@google.com} \\\AND
  Idan Szpektor \\
Google AI\\
\tt\small{szpektor@google.com} \\\And
  Jonathan Berant \thanks{\hspace{5px}Work done at Google AI.}\\
Tel Aviv University,\\
Allen Institute for AI\\
\tt\small{joberant@cs.tau.ac.il}\\}

\date{}

\begin{document} 
\maketitle 

\begin{abstract}


Sentence fusion is the task of joining several independent sentences into a single coherent text. Current datasets for sentence fusion are small and insufficient for training modern neural models. In this paper, we propose a method for automatically-generating fusion examples from raw text and present 
\discofuse{}, a large scale dataset for discourse-based sentence fusion. 
We author a set of rules for identifying a diverse set of discourse phenomena in raw text, and decomposing the text into two independent sentences.
We apply our approach on two document collections: Wikipedia and Sports articles, yielding 60 million fusion examples annotated with discourse information required to reconstruct the fused text. 
We develop a sequence-to-sequence model on \discofuse{} and thoroughly analyze its strengths and weaknesses with respect to the various discourse phenomena, using both automatic as well as human evaluation.
Finally, we conduct transfer learning experiments with \websplit{}, a recent dataset for text simplification. We show that pretraining on \discofuse{} substantially improves performance on \websplit{} when viewed as a sentence fusion task.
\end{abstract}

\section{Introduction}
\label{sec:intro}
Sentence fusion is the task of combining several independent sentences into a single coherent text \cite{barzilay2005sentence}. Sentence fusion is important in many NLP applications, including retrieval-based dialogue \cite{song2018ensemble, yan2018coupled}, text summarization \cite{barzilay2005sentence, Bing2015abstractive} and question answering \cite{li2018extraction, Marsi2005explorations}. Such systems retrieve multiple sentences from different sources, documents or paragraphs, and use them to construct a coherent text.

\begin{figure}
    \centering
    \includegraphics[scale=0.14]{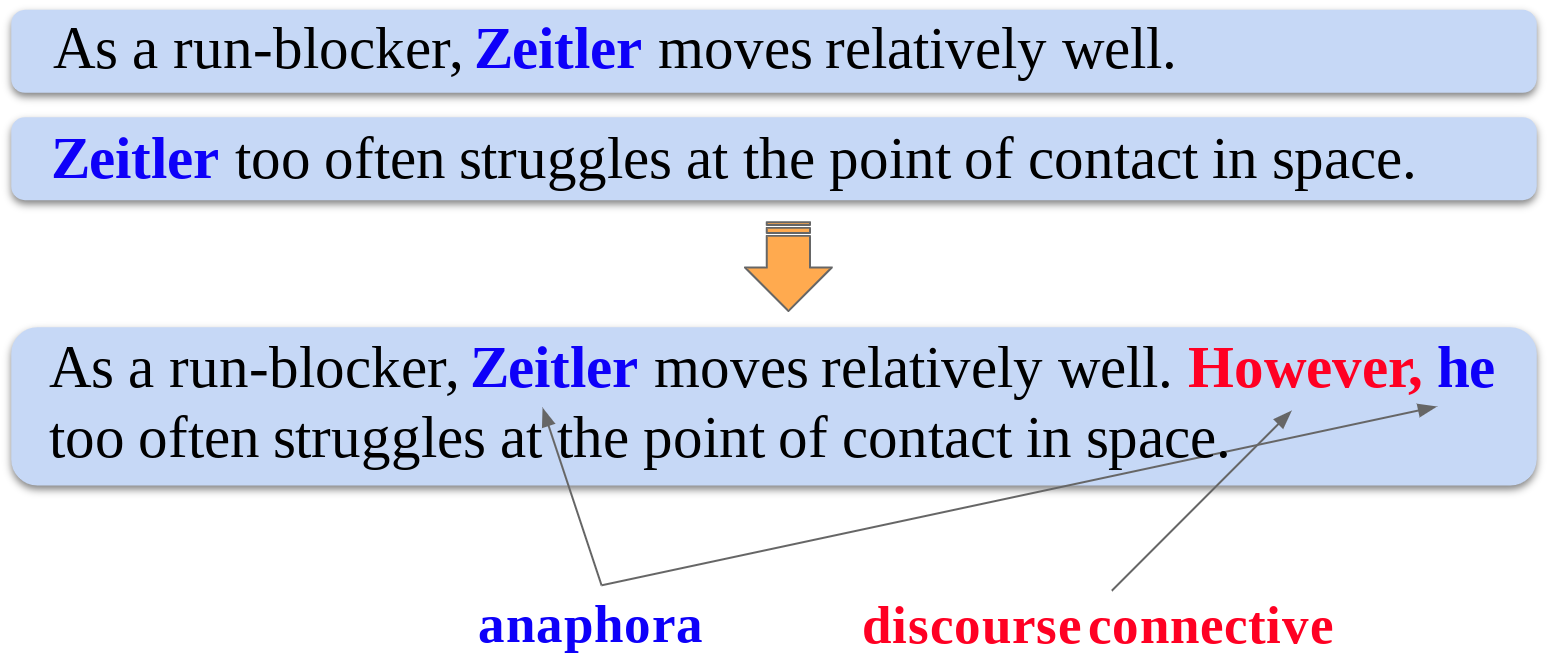}
    \caption{Example for two independent sentences, and their fusion. The modifications applied are pronominalization (\textcolor{blue}{blue}) and connective insertion (\textcolor{red}{red}).}
    \label{fig:first_page}
\end{figure}

Sentence fusion is challenging because it requires understanding the discourse semantics between the input sentences. Consider the example in Figure~\ref{fig:first_page}: a coherent fusion of the sentences requires understanding that the second sentence \emph{contrasts} the first one, in order to insert the discourse connective \nl{However}. In addition, the gender and syntactic role of the entity  \nl{Zeitler} needs to be inferred to insert the pronoun \nl{he}.

Prior work on sentence fusion \cite{barzilay2005sentence,turner2005supervised,filippova2010multi,elsner2011learning,thadani2013supervised,Bing2015abstractive,chali2017towards} utilized very small amounts of labeled data, which are insufficient to train modern neural models. In this work, we propose a method for automatically generating sentence fusion examples at scale from raw text corpora. 

To this end, we go over sentences and contiguous pairs of sentences in a corpus, and apply a set of manually-constructed rules, which identify the occurrence of prevalent fusion operations. The rules specify how to
modify the sentences such that they are ``unfused" into two independent
sentences. E.g., in Figure~\ref{fig:first_page} one rule will delete the discourse connective \nl{However}, and another will replace the pronoun \nl{he} with the named entity \nl{Zeitler}. 

In the generated examples, the original fused text becomes the target, and the unfused sentences (generated by rules) are the input.
Importantly, sentence fusion models trained on our data cannot simply learn to invert rule application, because information is lost and can be recovered only by understanding the text semantics . As mentioned, learning to insert \nl{However} in Figure~\ref{fig:first_page} requires inferring that the sentences contrast. We cover a wide range of fusion phenomena such as inserting discourse connectives in various positions of the sentences, anaphora and cataphora identification, and sentence merging through coordination, relative clauses and apposition.

We applied our method on two large document collections, Wikipedia and sports articles from the Web, resulting in two datasets of 16 million and 44 million examples respectively. We call the combined dataset \discofuse{}. We extensively analyze the quality of our dataset with crowdsourcing, and find that workers understand the text after splitting in 85\% of the cases, and the other 15\% are due to either the original text being unclear or errors in rule application.


\ignore{
We train baseline sequence-to-sequence models \cite{sutskever2009modelling,vaswani2017attention} and analyze the linguistic phenomena for which models struggle. 
Since there are many ways in which sentences can be fused, we augment automatic evaluation with human evaluation experiments, where crowdsourcing workers were asked to distinguish texts that were generated by our model and the original text.
We find that models trained on \discofuse{} generate human-like fusions overall, but raters' ability to detect model-generated fusions is statistically significant. 
Moreover, we find that structural modifications are easier to learn than semantic ones.
}

We trained a state-of-the-art sequence-to-sequence model \cite{vaswani2017attention} and analyzed the fusion phenomena in which the model struggles. 
We found that the model succeeds in fusing sentences through structural constructions such as apposition or relative clauses, but performs badly when fusion involves inserting a particular discourse connective, or selecting pronominals.

Last, we performed transfer learning by training on \discofuse{} and then fine-tuning on a smaller dataset from a different distribution. To this end, we utilize \websplit, a recent dataset for sentence splitting \cite{narayan2017split,aharoni2018split}, viewing \websplit as a sentence fusion task. 
We found that pre-training on \discofuse{} substantially improves the performance of a fusion model in this setup.

\ignore{
In addition, we perform an in-depth analysis using crowdsourcing, in order to both evaluate the models' performance as well as the quality of our dataset generation approach. Last, we evaluate our model on recent datasets suggested in the context of sentence splitting \cite{narayan2017split,aharoni2018split} to examine the extent to which training on a union-fusion dataset transfers to related tasks \mg{transfers to other data distributions?}.
}

To conclude, our contributions are:
\begin{enumerate}[topsep=0pt,itemsep=0pt,parsep=0pt,wide=0pt,leftmargin=\parindent]
\item \discofuse{}: a dataset of 60 million sentence fusion examples from two different corpora.
  \item A method for automatically generating sentence fusion examples from raw text.
  \item Automatic and human evaluation of the Transformer model on the fusion task.
  \item A transfer learning setting in which model performance improves when pre-trained with \discofuse{}.
\end{enumerate}
The \discofuse{} dataset is publicly available at: \url{https://discofuse.page.link/data}. 

\section{Background}
\label{sec:background}


Existing fusion datasets are small, which is perhaps why only few works have explored the application of supervised models to sentence fusion \cite{elsner2011learning,thadani2013supervised}. 
\newcite{mckeown2010time} introduced a
human-generated corpus of 3,000 examples.
\newcite{elsner2011learning} extracted around 300 fusion examples from pre- and post-editing news articles. 
\newcite{thadani2013supervised} constructed 1,858 examples from summarization tasks. Such datasets are too small to train modern data-hungry neural models. 

\ignore{
\jb{why is the next paragraph important? How does it related to us? Either
explain shortly the relation or delete?}
\citet{Chenal2016predicting} constructed the \textsc{SplitConj} dataset,
containing clusters of subsentential dependency trees that were split and
rearranged from 48,810 sentences in the Penn TreeBank. The goal of this dataset was to cluster dependency sub-trees that would fit for merging into a single sentence. 
}

Related to sentence fusion is its ``inverse'' task of sentence splitting.
\newcite{collados2013splitting} automatically constructed a Spanish simplification dataset by splitting single sentences into several simpler ones. 
Recently, two larger datasets for text splitting were released
\cite{botha2018learning, narayan2017split,aharoni2018split}.
However, using these datasets for the ``mirror'' task of sentence fusion is problematic. First, sentence splitting often involves removing content from the original sentence for simplification, and this content is impossible to recover in the fusion direction. Second, these datasets do not focus on discourse and thus prominent discourse phenomena may be missed. Last, our new dataset is more than an order of magnitude larger than the above sentence splitting datasets.



Another related line of recent work focused on predicting discourse
connectives between sentences and automatically generating examples from raw
text \cite{liu2016implicit,malmi2018automatic}. 
We substantially expand over those works by handling more diverse
linguistic phenomena, such as connectives in single sentences, generating anaphora and cataphora constructions, relative clauses, coordination and more, which are all represented in a single dataset. Moreover, our dataset is 20x larger compared to prior work, allowing us to examine in depth long-tail scenarios.

\section{The \discofuse{} dataset}

\begin{figure*}[th!]
    \centering
    \includegraphics[scale=0.24]{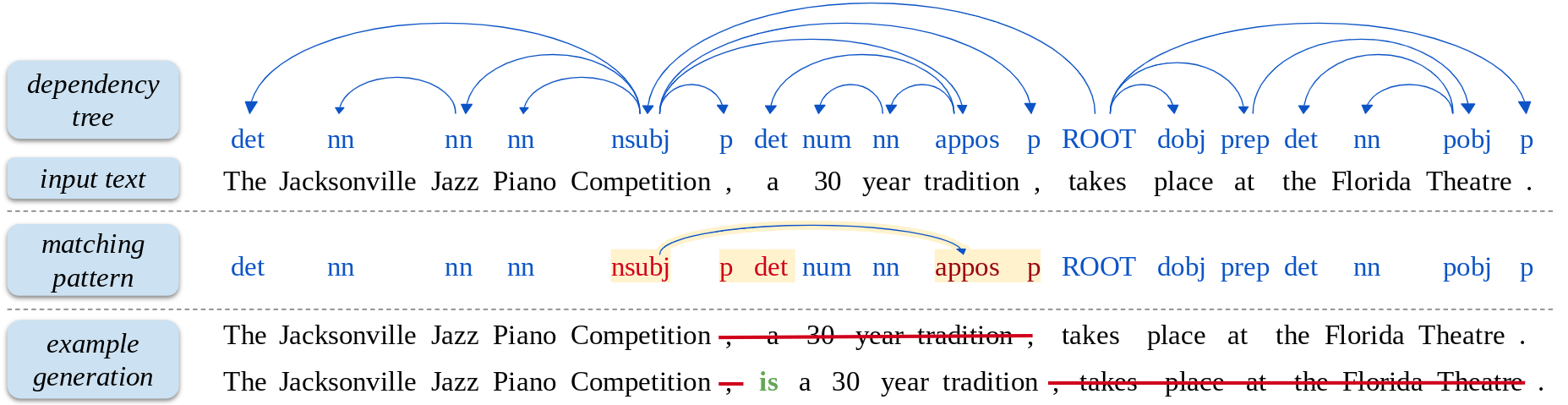}
    \caption{Example generation rule for apposition. Given an input text and its dependency tree, we check for a match with the apposition pattern. We then use the dependency tree to split the sentence and create a new example.}
    \label{fig:rule_illustration}
\end{figure*}

\begin{table*}[t]
\begin{center}
\setlength{\tabcolsep}{4pt}
\footnotesize{
\begin{tabular}{ll}
 \toprule
 \textbf{Phenomenon} & \textbf{Example} \\
 \hline
Discourse & \textcolor{blue}{(A)} Hebden Bridge is a popular place to live . \\
connective & \textcolor{blue}{(B)} \textbf{However ,} space is limited due to the steep valleys and lack of flat land . \\
 & \textcolor{red}{(a)} Hebden Bridge is a popular place to live . \\
 & \textcolor{red}{(b)} Space is limited due to the steep valleys and lack of flat land . \\
 \hline
 Anaphora & \textcolor{blue}{(A)} \textbf{Rider} entered the weekend averaging 23.0 points , good for 10th in the league . \\
 & \textcolor{blue}{(B)} \textbf{He} said those numbers mean little because of the Hawks ' 11 - 18 record. \\
 & \textcolor{red}{(a)} \textbf{Rider} entered the weekend averaging 23.0 points , good for 10th in the league . \\
 & \textcolor{red}{(b)} \textbf{Rider} said those numbers mean little because of the Hawks ' 11 - 18 record. \\
 \hline
Forward & \textcolor{blue}{(A)} \textbf{Although} the friendship somewhat healed years later , it was a devastating loss to Croly . \\
connective & \textcolor{red}{(a)} The friendship somewhat healed years later . \\
 & \textcolor{red}{(b)} It was a devastating loss to Croly . \\
 \hline
  Inner & \textcolor{blue}{(A)} Open workouts are held every Sunday \textbf{unless} the gym is closed for a holiday or other special events . \\
connective & \textcolor{red}{(a)} Open workouts are held every Sunday . \\
 & \textcolor{red}{(b)} The gym is closed for a holiday or other special events . \\
 \hline
   Cataphora & \textcolor{blue}{(A)} \textbf{Stating} that the proponents were unlikely to succeed in this appeal , \\
 & \hspace{1.4em} \textbf{Walker} rejected the stay request on October 23 . \\
 & \textcolor{red}{(a)} \textbf{Walker stated} that the proponents were unlikely to succeed in this appeal . \\
 & \textcolor{red}{(b)} \textbf{Walker} rejected the stay request on October 23 . \\
 \hline
 Sentence & \textcolor{blue}{(A)} The time of the autumn floods came , \textbf{and} the hundred streams poured into the Yellow River . \\
 coordination & \textcolor{red}{(a)} The time of the autumn floods came . \\
 & \textcolor{red}{(b)} The hundred streams poured into the Yellow River . \\
 \hline
 Verb phrase & \textcolor{blue}{(A)} The Sharks \textbf{started} the year 0 - 4 , \textbf{yet recovered} to claim sixth spot . \\
 coordination & \textcolor{red}{(a)} The Sharks \textbf{started} the year 0 - 4 . \\
 & \textcolor{red}{(b)} The Sharks \textbf{recovered} to claim sixth spot . \\
 \hline
  Relative & \textcolor{blue}{(A)} Kubler \textbf{, who} retired from cycling in 1957 \textbf{,} remained a revered figure in the wealthy alpine nation . \\
clause & \textcolor{red}{(a)} Kubler remained a revered figure in the wealthy alpine nation . \\
 & \textcolor{red}{(b)} Kubler retired from cycling in 1957 . \\
 \hline
Apposition & \textcolor{blue}{(A)} The frigidarium \textbf{,} the last stop in the bathhouse \textbf{,} was where guests would cool off in a large pool . \\
 & \textcolor{red}{(a)} The frigidarium was where guests would cool off in a large pool . \\
 & \textcolor{red}{(b)} The frigidarium is the last stop in the bathhouse . \\
\toprule
\end{tabular}}
\end{center}
  \caption{Generated fusion examples for different phenomena. The input text is marked in uppercase \textcolor{blue}{blue}, and the generated sentence pair is marked in lowercase \textcolor{red}{red}. We show in \textbf{boldface} parts that allow us to detect the target phenomenon.}
\label{tab:examples}
\end{table*}

We next describe our process for building \discofuse{}, which contains 60 million sentence fusion examples from two different document collections: Wikipedia and Web articles about sports.

\subsection{Example Generation}
\label{ssec:example_gen}

\discofuse{} contains union-fusion examples, \ie{} fusing sentences without loss of content \cite{Marsi2005explorations}. To automatically extract examples, we manually crafted a list of text splitting rules.
Our rule-set covers $9$ fusion phenomena, including handling discourse connectives, coordination and relative clauses, and entity resolution for anaphora and cataphora constructions. For entity resolution, both anaphoric pronouns (\nl{she}, \nl{they}, \nl{his}) and anaphoric nominals (\nl{the team}, \nl{the man}) are considered, based on the output of a coreference system.
The covered phenomena are summarized in Table~\ref{tab:examples} and a detailed description is given in Appendix~\ref{subsec:supplemental_rules}.

Given a text $t$ consisting of one or two consecutive sentences, each of our rules addresses a specific discourse phenomenon and has two parts: (a)~conditions for matching the phenomenon in $t$, and (b)~operations over a dependency tree annotated with coreference resolution. Applying the operations generates a fusion example $(x=(s_1, s_2), t)$, in which $(s_1, s_2)$ are two independent sentences originating from $t$, but stripped from the discourse phenomenon that tied them in $t$. 

Figure~\ref{fig:rule_illustration} gives an example of a rule for the apposition structure. The rule is applied to the sentence \nl{The Jacksonville Jazz Piano Competition, a 30 year tradition, takes place at the Florida Theatre}. First, the input is matched to the rule's condition. In this case, the condition is a single clause surrounded by two commas, which has a determiner as its first token and includes an apposition with incoming edge from a preceding token to the clause. Once matched, an example is generated. For this rule, the first sentence is created by removing the apposition clause, and the second sentence by removing the part after the clause and inserting the appropriate \nl{be} verb (\nl{is}). Generation examples for all 9 rule types are provided in Table~\ref{tab:examples}.


As explained in Section~\ref{sec:intro}, solving sentence fusion involves more than just reverse-engineering the generation rules. The model needs to decide whether to insert a discourse connective with the right semantics, whether to merge the input sentences, and what syntactic construction  (relative clause, coordination, apposition) is most appropriate in the given context.

Last, often several discourse phenomena occur in a single text $t$. Thus, we allow combining anaphora rules with one of the following rule types: discourse connective, inner connective and sentence coordination, which cover frequent combinations in our texts.

\subsection{Building the \discofuse{} Dataset}

To create \discofuse{} we retrieved the latest Wikipedia release and crawled the Web for several million sports articles. Documents were annotated with dependency trees and coreference resolution using Google Cloud Natural Language.\footnote{https://cloud.google.com/natural-language/}

We considered each sentence and pair of consecutive sentences in each document as candidates, applying the example generation process described in Section~\ref{ssec:example_gen}.
Additionally, we added as examples sentence pairs from the original corpus that did not match any rule, that is $(s_1, s_2) = t$, so that a trained model would also learn when not to change the input.
We filtered out examples with sentence length $\leq6$ tokens, and examples with non-ASCII characters.
\ignore{
reproducibility - the process we did is impossible to reproduce without code.
or special punctuation marks such as question mark and semi-colon. 
}
\ignore{
Finally, to adhere to legal requirements, from each Web sports document we randomly kept no more than 10 non-overlapping\footnote{Each sentence was allowed to appear in at most one example.} examples from each document. 
}
This process resulted in $44,177,443$ sports examples and $16,642,323$
Wikipedia examples. We randomly split these examples into 98\% train, 1\% dev, and 1\% test sets, making sure that each document contributes examples to only one of the split sets.


Like prior work \cite{malmi2018automatic}, we observed a skewed distribution of discourse phenomena in the data. Specifically, examples with anaphora or the connectives \nl{and} and \nl{but} constitute 99.7\% of Sports and 59\% of Wikipedia examples. 
Such a skewed distribution is likely to bias models and will 
fail to elucidate the ability of models to capture a wide range of linguistic
phenomena.
Therefore, we constructed a version of \discofuse{} by down-sampling examples containing \nl{and} and \nl{but} or anaphora. The down-sampled dataset contains 12,080,513 Sports examples and 4,581,352 Wikipedia examples.

The resulting distributions of discourse types and most common connectives in the two parts of \discofuse{} are provided in Appendix~\ref{subsec:data_distribution}. 
We will release both the original and the down-sampled versions of \discofuse{}.

\section{\discofuse{} Quality Evaluation}
\label{sec:quality}


\begin{table} [t]
\begin{center}
\footnotesize{
\begin{tabular}{c|r|r}
Rater Selection & \multicolumn{1}{c|}{\textsc{Sports} (\%)} & \multicolumn{1}{c}{\textsc{Wikipedia} (\%)} \\
\hline
Yes & 83.4 & 86.0 \\
\hline
No majority & 10.9 & 8.9 \\
\hline
No & 5.7 & 5.1 \\
\end{tabular}}
\end{center}
\caption{Rater evaluation understandability of the text after splitting. For each example, the majority of 5 raters was taken as the final rater selection.}
\label{tab:text_clarity}
\end{table}

\begin{table}[t]
\begin{center}
\setlength{\tabcolsep}{4pt}
\footnotesize{
\begin{tabular}{p{1.1cm}|p{6cm}}
Reason & \multicolumn{1}{c}{Example} \\
\hline
Original text &  \textcolor{blue}{(A)} UPDATE: Peat falls because footwork and quickness. \\
unclear & \textcolor{red}{(a)} UPDATE: Peat falls. \\
 & \textcolor{red}{(b)} Footwork and quickness. \\
\hline
Missing context &  \textcolor{blue}{(A)} We were right on the heels of Spurs, although Everton were closing in. \\
 & \textcolor{red}{(a)} We were right on the heels of Spurs. \\
 & \textcolor{red}{(b)} Everton were closing in. \\
\hline
Bad rule generation & \textcolor{blue}{(A)} He told reporters after the game his reaction was because he missed a wide-open Randall Cobb in the end zone. \\
 & \textcolor{red}{(a)} He told reporters after the game his reaction was. \\
 & \textcolor{red}{(b)} He missed a wide-open Randall Cobb in the end zone. \\
\end{tabular}}
\end{center}
\caption{Examples for three possible reasons for not understanding the text. In each example, \textcolor{blue}{(A)} is the original text and \textcolor{red}{(a)} and \textcolor{red}{(b)} are the two sentences generated by our rules.}
\label{tab:understandable}
\end{table}


\begin{table}[t]
\begin{center}
\footnotesize{
\begin{tabular}{p{1.7cm}|p{4.9cm}}
Error & \multicolumn{1}{c}{Generated sentence} \\
\hline
Extra comma & The space behind the fence in right field is blocked off \textcolor{red}{\textbf{,}} .  \\
\hline
Missing determiner & \textcolor{blue}{\textbf{My}} internist sent me for a mammogram and sonogram . \\
\hline
Bad pronoun replacement &  \textcolor{red}{\textbf{Lions}} have a 3 - 1 record overall against Lions   . \\
\end{tabular}}
\end{center}
\caption{Examples of grammatical errors introduced by our rules. The \textcolor{red}{red} text was incorrectly inserted and the \textcolor{blue}{blue} text was incorrectly removed.}
\label{tab:grammatical_errs}
\end{table}

To assess the quality of the generated fusion examples in \discofuse{}, we randomly selected 500 examples from each of the development sets of the Wikipedia and the Sports parts. We then conducted a crowdsourcing experiment in which each example was rated by 5 proficient English speakers,
limiting each rater to at most 6 items. Each rater was presented with the two independent sentences in the example and was asked to indicate whether the text is understandable. If the rater answered ``yes'', she was then asked to characterize the relation between the sentences and how she would fuse them. We next detail the results.

\subsection{Example Text Clarity}
\label{ssec:example_clarity}

Raters were asked whether they can understand the text after the example is split. Table~\ref{tab:text_clarity} summarizes this evaluation. Most examples were marked as understandable by the raters  (``yes'') -- 86\% of Wikipedia examples and 83.4\% of Sports examples. The rest either had no majority of rater votes or were marked as not understandable.  

To shed light on the possible reasons for obscurity, we analyzed 70 random examples that were not marked as understandable by the majority of raters. In 29 examples (41\%) the original text was unclear and for 17 examples a broader context was needed (24\%). In the remaining 24 examples (34\%), our rules generated sentences with grammatical or semantic errors. Examples for these cases are in Table~\ref{tab:understandable}.


Additionally, we analyzed 100 random examples for grammatical errors, and found that our rules did not introduce any errors in 79 examples. For 15 examples, the errors did not modify the meaning of the text nor caused semantic errors. The detected grammatical errors include extra commas, missing determiners and bad pronoun replacements, and are demonstrated in Table~\ref{tab:grammatical_errs}.

\subsection{Fusion Evaluation}

Next, we evaluated agreement on the fusion task for the 847 examples marked as understandable in Section~\ref{ssec:example_clarity}. 
Because there are many ways in which sentences can be fused, one cannot expect raters to produce the original text $t$ verbatim. 
Instead, we analyzed three central decisions and estimated whether people agree on those:
(a) whether to merge the two sentences into a single one or keep them separate; (b) whether there are entities in the text that should be replaced with nominal or pronominal anaphors or cataphors; and (c) which discourse connective to add (if any). 

For the last question, we presented raters with one connective from each of the four coarse-grained senses for discourse connectives defined by the PDTB \cite{prasad2008penn}:
\emph{comparison}, \emph{expansion}, \emph{contingency} and \emph{temporal}, as well as a \emph{no-connective} option. If the original text in the example includes a connective, we provided it as one of the options.

We observed a strong bias among raters towards refraining from performing any changes. E.g., while only 38\% of the examples did not contain a connective in $t$, the raters chose not to add a connective in 69.2\% of the cases. Similarly, only in 29.1\% of the examples the two sentences were not merged into a single one, while the raters chose not to merge in 53.1\% of the examples. Similar behavior was also observed by \newcite{malmi2018automatic} and  \newcite{rohde2016filling}.

We further looked at the agreement between the rater majority and the `gold' fusion decision. This analysis is shown in Table~\ref{tab:human_performance}. Agreement on merging the input sentences into one is almost random (52\%), since usually both options are valid. Consensus on whether to add an anaphor is higher, but not very high (63\%), especially in sentences when the anaphor in $t$ is a nominal rather than a pronoun. Finally, there is higher agreement on selecting the connective category (57\%), for which the random baseline is 20\%.

As mentioned, raters tend to keep the sentences unchanged. But in cases where raters agree to add a connective,
agreement figures increase substantially. Specifically, when it is clear that a connective is needed, there is also high agreement for picking the right one (76\%), for deciding whether to add an anaphor (70\%), and for deciding whether to merge the sentences or not (70\%).


\begin{table} [t]
\begin{center}
\footnotesize{
\begin{tabular}{|l|r|r|}
\hline
\multicolumn{1}{|c|}{Decision} & \multicolumn{1}{c|}{All Examples (\%)}  & \multicolumn{1}{c|}{Examples with} \\
\multicolumn{1}{|c|}{Type} & \multicolumn{1}{c|}{}  & \multicolumn{1}{c|}{connectives (\%)} \\
\hline
Single / pair & 52.0 & 70.1 \\
\hline
Anaphor & 63.4 & 70.1 \\
\hline
Connective & 57.0 & 76.4 \\
category & & \\
\hline
\hline
$\#$ Examples & 847 & 271 \\
\hline
\end{tabular}}
\end{center}
\caption{Average agreement for each fusion decision between the gold annotation and rater majority on examples marked as understandable by the raters. The right column considers only examples in which both the `gold' and rater majority agreed that a connective should be added.}
\label{tab:human_performance}
\end{table}

\section{Supervised Neural Fusion Models}
\label{sec:evaluation}

Using \discofuse{}, we trained a Transformer seq2seq model \cite{vaswani2017attention} that reads the input sentence pair and generates a fused text. We report model performance on the test-set using automatic metrics as well as human evaluation. We also provide detailed analysis of the different phenomena captured by this model.

\subsection{Experimental settings}
We tokenized all texts using byte-pair-encoding \cite{sennrich2015neural}
\ignore{
\footnote{https://github.com/google/sentencepiece} 
}
and compared the following three Transformer models : 
\begin{itemize}[topsep=4pt, itemsep=1pt, leftmargin=.2in, parsep=2pt]
    \item \textbf{\dfs{}} - trained on the sports portion of \discofuse{} after down-sampling.
    \item \textbf{\dfw{}} - trained on the Wikipedia portion of \discofuse{} after down-sampling.
    \item \textbf{\dfsw{}} - trained on a 50\%-50\% mixture of the sports and Wikipedia portions of \discofuse{} after down-sampling.
\end{itemize}

All models share the same network architecture, based on the best practices discussed by \citet{popel2018training}. We tuned parameters to select the best learning and dropout rates for each model with respect to the Exact Match objective (described in Section~\ref{ssec:automatic}). Network architecture and hyper-parameters are in Appendix~\ref{subsec:supplemental_neural}. As a baseline, we also tested a model called \textsc{Copy}, which simply concatenates the two input sentences.


\subsection{Automatic Evaluation Results}
\label{ssec:automatic}


\begin{table}[ht]
\begin{center}
\footnotesize{
\begin{tabular}{l|cccc|}
\textbf{\textsc{Sports}} & \multicolumn{2}{c}{Full} & \multicolumn{2}{c|}{Sampled} \\
& \sari{} & \exact{} & \sari{} & \exact{}{} \\
\hline
\dfs{} & \textbf{81.9} & \textbf{42.3}\% & \textbf{83.9} & \textbf{50.6\%} \\
\dfw{} & 77.8 & 31.7\% & 80.1 & 40.1\% \\
\dfsw{} & 80.7 & 38.3\% & 82.9 & 47.0\% \\
\hline
\textsc{Copy} & 40.0 & 1.1\% & 40.4 & 3.8\% \\
\multicolumn{4}{c}{}\\
\textbf{\textsc{Wikipedia}} & \multicolumn{2}{c}{Full} & \multicolumn{2}{c|}{Sampled} \\
& \sari{} & \exact{} & \sari{} & \exact{}{} \\
\hline
\dfs{} & 80.0 & 41.5\% & 80.0 & 41.9\% \\
\dfw{} & \textbf{83.1} & \textbf{47.6\%} & \textbf{84.5} & \textbf{51.1\%} \\
\dfsw{} & 82.8 & 46.7\% & 83.7 & 49.2\% \\
\hline
\textsc{Copy} & 40.3 & 1.0\% & 39.6 & 2.1\%
\end{tabular}}
\end{center}
\caption{\exact{} and \sari{} scores of \dfs{}, \dfw{}, \dfsw{} and \textsc{Copy}, on the test sets of \discofuse{} before (\emph{Full}) and after down-sampling (\emph{Sampled}).}
\label{tab:cross_domain}
\end{table}

\begin{table}
\begin{center}
\setlength{\tabcolsep}{4pt}
\footnotesize{
\begin{tabular}{p{3cm}|r|r|r|r}
\textbf{Discourse phenomena} & \multicolumn{2}{|c|}{\textbf{\dfs{}}} & \multicolumn{2}{c}{\textbf{\dfw{}}} \\
 & \exact{} & \sari{} & \exact{} & \sari{} \\
\hline
Apposition & 94.8\% & 99.3 & 94.7\% & 99.6 \\
Relative clause & 84.3\% & 95.3 & 76.9\% & 92.6 \\
Cataphora & 79.9\% & 92.8 & 84.2\% & 95.8 \\
\hline
Verb phrase coordination & 58.4\% & 88.0 & 58.7\% & 88.8 \\
None (control) & 55.3\% & 73.2 & 54.2\% & 72.7 \\
Anaphora & 52.1\% & 83.7 & 47.7\% & 81.5 \\
Inner connective & 49.9\% & 83.9 & 51.6\% & 85.5 \\
\hline
Sentence coordination & 35.6\% & 80.9 & 31.7\% & 79.4 \\
Inner connective + anaphora & 32.6\% & 82.5 & 37.2\% & 83.7 \\
Forward connective & 27.5\% & 80.2 & 34.6\% & 82.8 \\
Sentence coordination + anaphora & 20.5\% & 80.9 & 16.3\% & 78.3 \\
Discourse connective & 14.2\% & 65.6 & 29.1\% & 73.4 \\
Discourse connective + anaphora & 2.5\% & 73.0 & 8.0\% & 72.1 \\
\end{tabular}}
\end{center}
\caption{In-domain evaluation with breakdown by discourse phenomena. Performance of \dfs{} and \dfw{} on the sports and Wikipedia development sets. }
\label{tab:in_domain}
\end{table}

We evaluated model performance using two automatic metrics. The first is Exact Match (\exact{}) to see how often the model generates the exact same text as the gold fusion. The second is \sari{} \cite{xu2016optimizing}, which computes the set of added, removed, and kept n-grams in the model output, comparing the output both with the gold text and the input text. Then it computes the F$_1$ scores for these three sets and averages the scores. We compute \sari{} on up to 4-grams, as in \citet{xu2016optimizing}.
We refrained from using metrics like BLEU because in fusion there is large overlap between the input sentences and their fused version, and such metrics do not capture well fine-grained differences of only a single word.

We note that our definition of \sari{}\footnote{Our \sari{} implementation is available at: \url{https://github.com/tensorflow/tensor2tensor/blob/master/tensor2tensor/utils/sari_hook.py}} slightly differs from the one given by \newcite{xu2016optimizing} in two aspects: ($i$)~
We define $\frac{0}{0}=1$ when computing precision and recall,
otherwise \sari{} could be less than 1 even if the output matches the gold text exactly. ($ii$)~Instead of considering only the precision of deleted n-grams, we use F$_1$ for all three sets. Otherwise, \sari{} will give high scores to models that merely copy everything in the input, without even trying to infer what to delete.


Table~\ref{tab:cross_domain} summarizes the results. When training and testing on the same domain, either Sports or Wikipedia, \sari{} score is a little above 80 points for the full dataset. Yet \exact{} is not high, around 42\% for Sports and 47\% for Wikipedia, showing that in the majority of the examples the model's fusion differs from the gold. Tested on the down-sampled test-set, performance increases significantly for \exact{}, especially on Sports, where discourse phenomena is more skewed.

We next turn to cross-domain evaluation. When applying a model trained on one domain to the other domain performance drops. This shows that the discourse phenomena distribution differs between the domains, indicating that transfer learning is not trivial even with these large datasets. This is especially evident when applying \dfw{} to Sports, where \exact{} falls from 42\% to 32\% on the full dataset and from 50\%  to 40\% on the down-sampled one. Interestingly, when learning on the mixed training set, performance on both domains is close to in-domain performance, showing that the model has the capacity to handle both domains.

Finally, we take advantage of the provided annotation of the different discourse phenomena within each example in \discofuse{}. We conducted a detailed analysis of in-domain model performance by discourse type, presented in Table~\ref{tab:in_domain}. Results show that structural discourse types, such as apposition and relative clause, are easier to learn with both high exact match and \sari{} scores. While differences with respect to \sari{} scores are not large between phenomena, exact match varies more. Anaphora and verb phrase coordination are more challenging, but still require matching of the same noun (the named entity or the subject). On the other hand, discourse types that involve connective prediction, such as sentence coordination and discourse connective, require semantic understanding, and performance is significantly lower. In addition, when two discourse types are required for fusion, performance drops dramatically.



\subsection{Human Evaluation Results}

\begin{table}
\setlength{\tabcolsep}{5pt}
\begin{center}
\footnotesize{
\begin{tabular}{ll|rr}
 & & $\#$ Examples & Detection \\
\hline
& output = gold & 525 & 50\% \\
\textsc{Sports} & output != gold & 475 & 65\% \\
& total & 1000 & \textbf{57\%} \\
\hline
& output = gold & 528 & 50\% \\
\textsc{Wikipedia} & output != gold & 472 & 61\% \\
& total & 1000 & \textbf{55\%} \\
\end{tabular}}
\end{center}
\caption{Human detection (\emph{Detection}) percentage for \dfs{} and \dfw{} on 1000 samples from each of the Sports and Wikipedia development sets. We report \emph{Detection} for cases when model output differed from the gold, and cases when they were identical.}
\label{tab:human_test}
\end{table}

As our second experiment we employed crowd-sourcing to test how distinguishable the fusion model outputs are from the gold fused texts. Concretely, we present raters an independent sentence pair from \discofuse{} and two fused versions - the gold version and one generated by a model. Raters were asked to detect the gold version. For each example, we took the majority of $5$ raters as the final choice. This experiment mitigates the difficulties of automatic text generation evaluation, where many outputs are valid for a single input.

We sampled $1000$ random examples from each development set of the two domains and applied the in-domain model to both. The raters were presented only with examples where the model output was different from the gold fusion, and assumed 50\% detection accuracy otherwise. 

Table~\ref{tab:human_test} depicts the results. 
Out of cases when model output differed from the gold, raters were able to identify the human version in 65\% of Sports examples and 61\% of Wikipedia examples. Looking at the entire set, humans were able to identify the human version in 57\% (Sports) and 55\% (Wikipedia) of the cases.
This shows that our Transformer model, applied over a dataset of millions of examples, is able to learn good fusions in general. Nevertheless, models are still far from perfect -- human accuracy is clearly better than random and this improvement is statistically significant at a level of $p < 10^{-5}$ for Sports and $p < 10^{-3}$ for Wikipedia. 
\ignore{
discourse type & accuracy \\
PAIR\_ANAPHORA & 0.559 \\
PAIR\_CONN & 0.641 \\
PAIR\_CONN\_ANAPHORA & 0.619 \\
PAIR\_NONE & 0.586 \\
SINGLE\_APPOSITION & 0.556 \\
SINGLE\_CATAPHORA & 0.647 \\
SINGLE\_CONN\_INNER & 0.640 \\
SINGLE\_CONN\_INNER\_ANAPHORA & 0.582 \\
SINGLE\_CONN\_START & 0.686 \\
SINGLE\_RELATIVE & 0.609 \\
SINGLE\_S\_COORD & 0.588 \\
SINGLE\_S\_COORD\_ANAPHORA & 0.694 \\
SINGLE\_VP\_COORD & 0.594
}


\subsection{Alignment-based Analysis}

We next present an analysis of the types of errors our models produce. To this end, we sampled 40K examples of \dfs{} and \dfw{} outputs on Sports and Wikipedia development sets. We then automatically aligned predicted sequences to gold sequences and looked at the differences between aligned words. The trained models successfully learned to copy most of the input text, and thus errors due to alignment problems are rare.

We start by considering the semantic relation between the input sentences.  Table~\ref{tab:connective_prediction} displays model accuracy in predicting the most common connectives in \discofuse{}, as well as the top connectives predicted in this slot. 
We observe that when the model predicts a wrong connective, that connective is often reasonable, e.g., predicting \nl{but} instead of \nl{and} or \nl{however}.
Moreover, a second source of error is not adding a connective at all. It is also clear that some connectives, like \nl{however}, \nl{although} and \nl{for example}, are harder to learn.

\begin{figure}[t]
\centering
\includegraphics[scale=0.23]{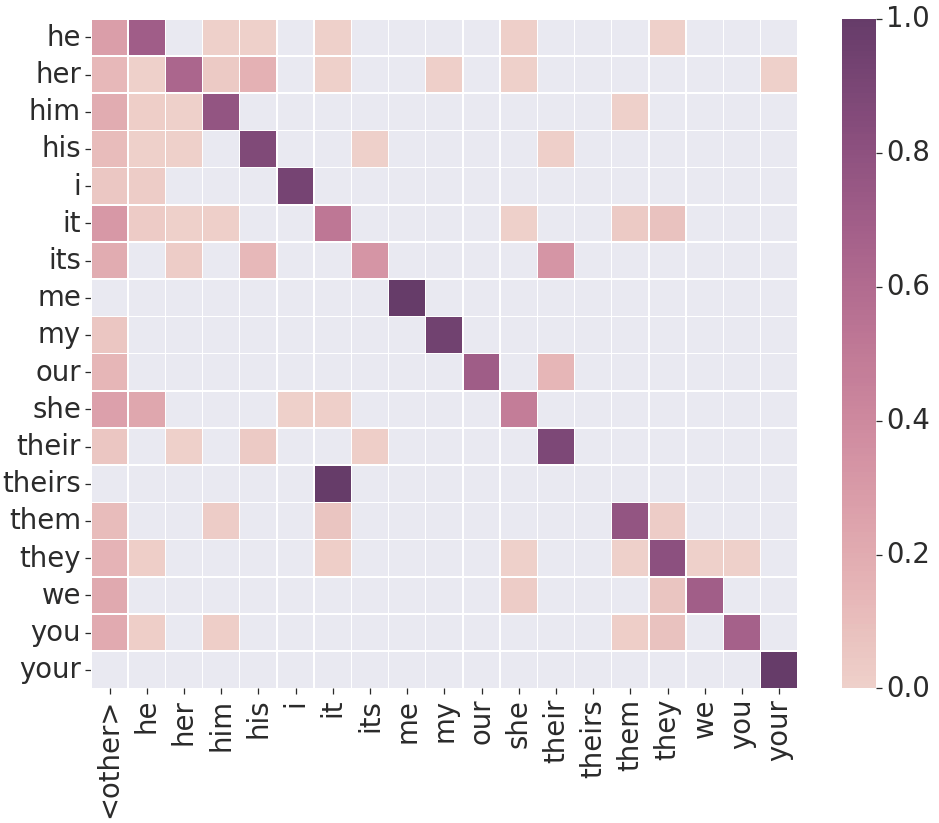}
\caption{\dfw{} outputs versus the gold pronouns. Rows refer to gold pronouns and columns refer to aligned model outputs at the gold pronoun position. Values in each row are normalized to 1. Column \angleb{other} refers to model outputs that are not pronouns.}
\label{fig:pronoun_matrix}
\end{figure}

We also analyzed the models' ability to correctly infer pronoun anaphors including gender, possessive and plurality.
Figure~\ref{fig:pronoun_matrix} shows the pronoun confusion matrix for \dfw{},\footnote{Results for \dfs{} are very similar.} where lines refer to gold pronouns and columns to the generated pronoun in the same position. The clear diagonal shows that in most cases, the model successfully outputs the correct pronoun. However, the \angleb{other} column indicates that occasionally the model tends not to replace the entity in the input with a pronoun anaphor. In addition, the model seems to struggle with possession and plural 3rd person (\nl{it}, \nl{its}, \nl{they}, \nl{their}, \nl{theirs}).



\begin{table}
\begin{center}
\footnotesize{
\begin{tabular}{p{1.3cm}|c|c|p{2.26cm}}
\toprule
Connective & \dfs{} & \dfw{} & Top 3 connectives \\
 & accuracy &  accuracy &  \\
\hline
and & 50.9 & 53.7 & and, but, \angleb{other} \\
\hline
but & 42.8 & 43.7 & but, \angleb{other}, and \\
\hline
because & 61.5 & 60.7 & because, \angleb{other}, but \\
\hline
although & 35.1 & 33.2 & although, \angleb{other}, but \\
\hline
so & 50.6 & 50.2 & so, but, and \\
\hline
or & 70.5 & 72.1 & or, and, \angleb{other} \\
\hline
however & 28.3 & 26.7 & \angleb{other}, however, but \\
\hline
while & 70.1 & 70.6 & while, \angleb{other}, but \\
\hline
so that & 64.3 & 63.0 & so that, \angleb{other}, because \\
\hline
unless & 68.9 & 67.0 & unless, because, \angleb{other} \\
\hline
for example & 26.9 & 28.1 & \angleb{other}, for example, however \\
\toprule
\end{tabular}}
\end{center}
\caption{Alignment-based connective prediction accuracy for the most common connectives. When a model did not add a connective, the token \angleb{other} is used.}
\label{tab:connective_prediction}
\end{table}

\section{Transfer Learning Experiment}


With the \discofuse{} approach we can collect a large amount of examples automatically. Still, these examples only reflect the manual rules that identify discourse phenomena. We wanted to see if \discofuse{} covers enough cases such that a trained model would be helpful for testing on fusion datasets generated by different approaches.

\subsection{Experimental settings}

In this experiment, we looked at the recently released \websplit{} dataset 1.0~\cite{narayan2017split}. It consists of examples $(t, \{s_i\}_{i=1}^n)$, where $t$ is a sentence that verbalizes the same set of RDF triples as $\{s_i\}_{i=1}^n$. We note that \websplit{} was originally developed for sentence splitting, from $t$ to $\{s_i\}_{i=1}^n$, but here we view its examples for the reverse fusion task: from $\{s_i\}_{i=1}^n$ to $t$. We only considered examples where $\{s_i\}_{i=1}^n$ corresponds to exactly two simpler sentences ($n=2$). This leaves us with 135K training, 8K validation, and 8K test samples.

We tokenized the data using byte-pair-encoding \cite{sennrich2015neural} and compared three models: ($i$)~The \textsc{Copy} baseline that concatenates the two input sentences, ($ii$)~a model trained on \websplit{} alone, and ($iii$)~a model pre-trained on \dfw{} and fine-tuned on \websplit{}. 

For the last two models, we use the CopyNet architecture \cite{gu2016copying}, which is similar to state-of-the-art models for the splitting task on \websplit{}~\cite{narayan2017split,botha2018learning}. While the Transformer outperformed this model on our main experiments, here it overfit on the small training set of \websplit{}. The training details are provided in Appendix~\ref{subsec:supplemental_neural}.




\subsection{Results}

\begin{table}
\begin{center}
\footnotesize{
\resizebox{\columnwidth}{!}{%
\begin{tabular}{lcccc}
\toprule
Training data & \sari{} & Keep & Add & Delete  \\
\midrule
\textsc{Copy}       &  18.1 & 52.9 &  0.5    &    0.9 \\
\websplit        &  40.5 & 44.6 &  7.8    &   \textbf{69.3} \\
\dfw{} + \websplit   &  \textbf{44.2} & \textbf{54.8} & \textbf{10.4} & 67.5 \\
\bottomrule
\end{tabular}
}}
\end{center}
\caption{Fusion results on \websplit{}, measured by \sari{} and the F1 scores that compose it.
}
\label{tab:transfer}
\end{table}

Table~\ref{tab:transfer} shows the results of the experiment. Similarly to Section~\ref{sec:evaluation}, we measured the model performance using \sari{}.
Pre-training with \dfw{} improves \sari{} score by 9\% compared to using \websplit{} alone. In particular, the F1 of the `kept' and `added' n-grams is significantly higher, by 23\% and 33\% respectively. Specifically, `added' tokens refer also to correctly choosing discourse connectives, to which the large-scale examples in \discofuse{} were likely helpful. 

We note that even with pre-training, the SARI `add' score is only 10.4. This is probably due to the large amount of paraphrasing done in \websplit{}, which makes it problematic for fusion evaluation (see also Section~\ref{sec:background}). For example:
\begin{itemize}[topsep=5pt, itemsep=5pt, leftmargin=.2in, parsep=2pt]
    \small
    \item[] \textbf{Sentence 1:} \textit{Bolt , a comic character AKA Larry Bolatinsky , was created by Paris Cullins and Ernie Colon .}
    \item[] \textbf{Sentence 2:} \textit{Paris Cullins is a United States national .}
    \item[] \textbf{Gold:} \textit{Larry Bolatinsky is \textcolor{red}{the alternative name for the} comic \textcolor{red}{book} character Bolt , \textcolor{red}{which} was created by Ernie Colon and \textcolor{red}{the American} Paris Cullins .}
\end{itemize}
Correctly inferring the added terms (shown in red) requires paraphrasing knowledge that is outside the scope of \discofuse{}.


\section{Conclusions}
We presented \discofuse{}, a large-scale dataset for sentence fusion that was generated by applying a rule-based method.
It contains millions of examples from two domains, annotated with multiple discourse phenomena.

We used \discofuse{} to build supervised neural models for sentence fusion and conducted fine-grained analyses of the results. 
Currently, our models fuse only two sentences together. We would like to expand them to more input sentences in future work.

We also demonstrated \discofuse{}'s usefulness in a transfer learning setup on a different fusion test-set, hoping it would facilitate research on text fusion in data-scarce domains.



\section*{Acknowledgments}
We thank Marta Recasens, Jan Botha, and Sascha Rothe from Google AI for helpful discussions. 
This research was supported by the Yandex Initiative in Machine Learning and by The Israel Science Foundation grant 942/16.
This work was completed in partial fulfillment for the Ph.D degree of the first author.

\bibliography{all}
\bibliographystyle{acl_natbib}

\clearpage
\appendix

\section{Supplemental Material}
\label{sec:supplemental}

\subsection{Generation Rules}
\label{subsec:supplemental_rules}
In this section we provide technical details of the generation rules used to create \discofuse{}.
For the sake of clarity, we provide a simplified version of the rules, that does not include edge cases and minor implementation details.
The discourse connectives we considered in the rules were selected from the Penn Discourse Treebank (PDTB) \cite{prasad2008penn} and are listed in Table~\ref{tab:sets}.

Given an input text, it is encoded with 3 lists: $Z$ is the token list, $Z_t$ is a list of POS tags, $Z_l$ is a list of dependency labels (see Table~\ref{tab:definitions}). In addition, all entities mentioned in the text are extracted and stored such that for two token lists $Z,R$, the set $m(Z,R)$ holds all the mention pairs of the same entity in the two lists. 
Each rule is designed for a specific discourse phenomenon and contains two parts. First, a set of conditions is applied to the input lists to detect whether the phenomenon occurs in the text. If a discourse pattern has been identified, a short sequence of simple operations is applied to the input, yielding a new sentence pair. Table~\ref{tab:operations} summarizes the operations in use, which allow insertion and deletion of tokens and splitting of the input text.

Table~\ref{tab:rules} provides the technicalities of each rule, i.e. the detection conditions of the discourse structure, and the sequence of operations for generating a new sentence pair from it. A detailed example for two-rule execution process is given in Table~\ref{tab:rule_example}.

As mentioned, the rules are simplified for clarity. However, we note two special cases where morphological modifications are required to produce text without grammatical errors. First, in some cases of forward connective and cataphora, the tense change of a verb is required when splitting the input sentence. For instance, in the cataphora example in Table~\ref{tab:examples}, we change the verb \nl{stating} to have a past tense -- \nl{stated}. Likewise, occasionally a \nl{be} verb needs to be inserted when splitting a single sentence, as demonstrated in Figure~\ref{fig:rule_illustration}. In our rules, we choose which \nl{be} verb to insert based on the tense and perspective of the rest of the sentence.

\begin{table}[ht!]
\begin{center}
\scriptsize{
\begin{tabular}{|lp{5.5cm}|}
\multicolumn{1}{l}{\textbf{Notation}} & \multicolumn{1}{p{5.5cm}}{\textbf{Definition}} \\
\toprule
\hline
Token list $Z$& A list of tokens $\{z^{(1)}, ..., z^{(|Z|)}\}$ \\
\hline
$Z_t$ & The list of POS tags of $Z$, where $Z_t^{(i)}$ is the tag of $z^{(i)}$ for every $i = 1, ..., |Z|$. \\
\hline
$Z_l$ & The list of dependency labels of $Z$, where $Z_l^{(i)}$ is the label of incoming edge of $z^{(i)}$ for every $i = 1, ..., |Z|$. \\
\hline
$m(Z,R)$ & A set of mention pairs in $Z,R$ : $\{ \langle S_Z ,\, S_R \rangle \mid S_R\prec R \text{ and } S_Z\prec Z \text{ are mentions of the same entity} \} $ \\
\hline\hline
$S \prec Z$ & $S$ is a span in $Z$, such that $\exists i \in 1, ..., |Z|-|S|+1: \forall j = 0, ..., |S|-1 : s_{1+j} = z_{i+j}$  \\
\hline
$S \sqsubset Z$ & $S$ is a prefix of $Z$, such that $\forall i = 1, ..., |S|: s_i = z_i$  \\
\hline
$i \curvearrowright_U j$ & There is an edge from the $i$th token to the $j$th token in the dependency tree of $Z$. \\
\hline\hline
$\mathcal{C}_b$ & A set of backward connectives.  \\
\hline
$\mathcal{C}_s$ & A set of intra-sentence connectives, which are either forward connectives or conjunctions. \\
\hline
$\mathcal{C}_f$ & A set of forward connectives. \\
\hline
$\mathcal{C}_c$ & A set of coordinating conjunctions. \\
\hline
$\mathcal{P}_r$ & A set of relative pronouns. \\
\hline
$\mathcal{V}$ & A set of POS tags for verbal phrases. \\
\hline
\end{tabular}}
\end{center}
\caption{Notation and definitions for Table \ref{tab:rules} of generation rules. $Z,R,S_Z,S_R$ are token lists and $1\leq i,j \leq |Z|$ are indices. The full lists of connectives and POS tags are provided in Table \ref{tab:connectives}.}
\label{tab:definitions}
\end{table}

\begin{table}[ht!]
\begin{center}
\scriptsize{
\begin{tabular}{lp{6.5cm}}
\textbf{Set} & \textbf{Values} \\
\hline
$\mathcal{C}_b$ & "accordingly", "additionally", "afterward", "alternatively", "although ,", "and", "as a result ,", "because of that", "because of this", "besides ,", "but", "by comparison ,", "by contrast ,", "by doing this ,", "by then", "consequently", "conversely", "else,", "finally ,", "for example", "for instance", "further ,", "furthermore", "hence ,", "however", "in contrast ,", "in fact ,", "in other words", "in particular ,", "in short ,", "in sum ,", "in the end ,", "in turn ,", "indeed ,", "instead ,", "lest", "likewise ,", "meantime ,", "in the meantime ,", "meanwhile ,", "moreover", "nevertheless", "next ,", "nonetheless", "on the contrary ,", "on the other hand", "or ,", "otherwise ,", "overall ,", "plus ,", "rather ,", "regardless ,", "similarly ,", "simultaneously", "specifically ,", "still ,", "then ,", "thereafter ,", "thereby ,", "therefore", "though ,", "thus ,", "ultimately ,", "whereas", "yet ,", "now ,", "second ,", "third ,", "basically ,", "this ,", "eventually ,", "obviously ,", "again ,", "fortunately ,", "luckily ,", "meaning ,", "interestingly ,", "anyway ,", "clearly ,"  \\
\hline
$\mathcal{C}_s$ & "because", ", because", "hence", ", while", "whereas", ", although", "although", "and although", "unless", "now that", ", now that", "so that", ", so that", "meaning", ", meaning" \\
\hline
$\mathcal{C}_f$ & although, since, in addition to, aside from \\
\hline
$\mathcal{C}_c$ & and, but, or, nor, yet, so, for \\
\hline
$\mathcal{P}_r$ & who, which, whose, whom \\
\hline
$\mathcal{V}$ & VB, VBD, VBG, VBN, VBP, VBZ \\
\hline
\end{tabular}}
\end{center}
\caption{Connectives and POS tags used in our detection rules. A preceding comma is allowed for conjunctions in $\mathcal{C}_c$. For the connectives \nl{although} and \nl{since} in $\mathcal{C}_f$, we do not allow a following comma.}
\label{tab:sets}
\end{table}

\begin{table}[ht!]
\begin{center}
\scriptsize{
\begin{tabular}{l|p{4.6cm}}
\textbf{Operation} & \textbf{Description} \\
\hline
$\textsc{Delete}(X,i,n)$ & Delete a sequence of $n$ tokens from $X$, starting from index $i$. \\
$\textsc{Prepend}(X,Y)$ & Attach the list $Y$ at the beginning of $X$. \\
$\textsc{Replace}(X,Y,Z)$ & Replace every occurrence of $Y$ in $X$ with $Z$, in a non-overlapping manner. \\
$\textsc{Split}(X,i)$ & Split $X$ into two token lists $V=\{x_1,...,x_{i-1}\}, W=\{x_i,...,x_{|X|}\}$. \\
$\textsc{Trim}(X)$ & Delete all tokens in $X$ after the first punctuation token, e.g. period, comma, etc. \\
\hline
\end{tabular}}
\end{center}
\caption{Operations upon token lists, which are used for generation of sentence pairs (Table \ref{tab:rules}). The arguments $X,Y,Z$ are token lists and the arguments $i,n$ are integers.}
\label{tab:operations}
\end{table}

\begin{table*}[th!]
\begin{center}
\scriptsize{
\begin{tabular}{l|p{0.7cm}|l|l}
\toprule
\textbf{Phenomenon} & \textbf{Input} & \textbf{Detection} & \textbf{Generation} \\
\hline
Discourse connective & (A,B) & $\exists S \in \mathcal{C}_b \,,\, i\in [1,5] : S\sqsubset \{b^{(i)}, ..., b^{(|B|)}\} $ & $ \textsc{Delete}(B,i,|S|) $ \\
\hline
Anaphora & (A,B) & $ m(B,A) \neq \emptyset $ & $\textsc{Replace}(B, S_B, S_A)$ \\
& & &  $\;\forall \langle S_B ,\, S_A \rangle \in m(B,A) $  \\
\hline
\multirow{3}{*}{Forward connective} & \multirow{3}{*}{Z} & $\exists S \in \mathcal{C}_f : S\sqsubset Z $ & $A,B \leftarrow \textsc{Split}(Z, i) $ \\ & & $\exists i: |S|+1 < i < |Z| \wedge z^{(i)} = ","$ & $ \textsc{Delete}(A,1,|S|)$ \\ & & & $ \textsc{Delete}(B,1,1)$ \\
\hline
\multirow{3}{*}{Inner connective} & \multirow{3}{*}{Z} & \multirow{3}{*}{$\exists S \in \mathcal{C}_s : S\prec Z $} & $A,B \leftarrow \textsc{Split}(Z, i) $ \\ & & & $\textsc{Delete}(B,1,|S|)$ \\ & & & $ \textsc{Trim}(B) $ \\
\hline
\multirow{3}{*}{Cataphora} & \multirow{3}{*}{Z} & $\exists i: 1 < i < |Z| \wedge z^{(i)} = ","$ & $A,B \leftarrow \textsc{Split}(Z,i) $ \\ & & $Z_t^{(1)} = \text{VBG} \wedge Z_l^{(1)} = \text{vmod} \wedge Z_t^{(i+2)} \in \mathcal{V} $ & $ \textsc{Prepend}(A,\{b_1\}) $ \\ & & & $ \textsc{Delete}(B,1,1)$ \\
\hline
\multirow{2}{*}{Sentence coordination} & \multirow{2}{*}{Z} & $\exists i<j \leq i+5 : Z_l^{(i)} = \text{cc} \,\wedge\, Z_l^{(j)} = \text{conj}$ & $A,B \leftarrow \textsc{Split}(Z,i) $ \\ & & $\exists i<k<j : Z_l^{(k)} \in \{\text{nsubj}, \text{nsubjpass}\} $ & $ \textsc{Delete}(B,1,1)$ \\ 
\hline
\multirow{3}{*}{Verb phrase coordination} & \multirow{3}{*}{Z} & $\exists i,j:i<j \leq i+5 : Z_l^{(i)} = \text{cc} \,\wedge\, Z_l^{(j)} = \text{conj} \,\wedge\, Z_t^{(j)} \in \mathcal{V}$ & $A,B \leftarrow \textsc{Split}(Z,i) $ \\ & & $\exists k : k\curvearrowright_Z j \,\wedge\, Z_l^{(k)} = \text{root} $ & $ \textsc{Delete}(B,1,1)$ \\ 
& & & $ \textsc{Prepend}(B,\{a_1, ..., a_{k-1}\}) $ \\
\hline
\multirow{3}{*}{Relative clause} & \multirow{3}{*}{Z} & $\exists i,j: 1 < i < j < |Z| \wedge z^{(i)} = z^{(j)} = ","$ & $A,B \leftarrow \textsc{Split}(Z,j) $ \\ & & $z^{(i+1)} \in \mathcal{P}_r$ & $ \textsc{Delete}(B,1,1)$ \\ 
& & & $ \textsc{Delete}(A,i,2)$ \\
& & & $\textsc{Prepend}(B,\{a_r, ..., a_{i-1}\})^\mathbf{*}$ \\
\hline
\multirow{3}{*}{Apposition} & \multirow{3}{*}{Z} & $\exists i,j: 1 < i < j < |Z| \wedge z^{(i)} = z^{(j)} = ","$ & $A,B \leftarrow \textsc{Split}(Z,j) $ \\ & & $Z_l^{(i+1)} \in \{\text{det, poss}\} \,\wedge\, \exists k: i<k<j, Z_l^{(k)} = \text{appos}$ & $ \textsc{Delete}(B,1,1)$ \\ 
& & & $ \textsc{Delete}(A,i,2)$ \\
& & & $\textsc{Prepend}(B,\{a_r, ..., a_{i-1}\})^\mathbf{*}$ \\
\toprule
\end{tabular}}
\end{center}
\caption{Generation rules for sentence pairs. The rules apply for token lists $Z,A,B$, where $Z$ represents a single sentence and $A,B$ either represent two consecutive sentences or two consecutive sentence parts. $^\mathbf{*}$For the rules of relative clause and apposition, $r$ is the index of the leftmost child in the dependency sub-tree of $a^{(i)}$.}
\label{tab:rules}
\end{table*}

\begin{table}[ht!]
\begin{center}
\scriptsize{
\begin{tabular}{p{6.7cm}p{0.1cm}}
\hline
\multicolumn{2}{|l|}{\textbf{1. Input}} \\
\multicolumn{2}{|p{6.8cm}|}{$Z = $ \{\textcolor{red}{Ruiz} ordered his first shot to be retaken \textcolor{red}{because} Brazilian players entered the penalty area before \textcolor{red}{his} kick .\}} \\

\hline \multicolumn{2}{p{6.8cm}}{\centering{\normalsize{\rotatebox[origin=c]{-90}{\MVRightarrow}}}} \\
\hline
\multicolumn{2}{|p{6.8cm}|}{\textbf{2. Inner connective}} \\
\multicolumn{2}{|p{6.8cm}|}{\emph{Detection}} \\
\multicolumn{2}{|p{6.8cm}|}{For $S = $ "because" it holds that $S \in \mathcal{C}_s$ and $S\prec Z $.} \\
\multicolumn{2}{|p{6.8cm}|}{} \\
\multicolumn{2}{|p{6.8cm}|}{\emph{Generation}} \\
\multicolumn{1}{|p{2cm}|}{$\textsc{Split}(Z, i) $} & \multicolumn{1}{p{4.8cm}|}{$A = $ \{\textcolor{red}{Ruiz} ordered his first shot to be retaken .\}} \\
\multicolumn{1}{|p{2cm}|}{} & \multicolumn{1}{p{4.8cm}|}{$B = $ \{\textcolor{red}{Because} Brazilian players entered the penalty area before \textcolor{red}{his} kick .\}} \\
\multicolumn{1}{|p{2cm}|}{$\textsc{Delete}(B,1,|S|)$} & \multicolumn{1}{p{4.8cm}|}{$B = $ \{ Brazilian players entered the penalty area before \textcolor{red}{his} kick .\}} \\
\multicolumn{1}{|p{2cm}|}{\textsc{Trim}(B)} & \multicolumn{1}{p{4.8cm}|}{\emph{no effect at this case}} \\

\hline \multicolumn{2}{p{6.8cm}}{\centering{\normalsize{\rotatebox[origin=c]{-90}{\MVRightarrow}}}} \\
\hline 
\multicolumn{2}{|p{6.8cm}|}{\textbf{3. Anaphora}} \\
\multicolumn{2}{|p{6.8cm}|}{\emph{Detection}} \\
\multicolumn{2}{|p{6.8cm}|}{$m(B,A) = \{(\text{his}, \text{Ruiz}) \} \neq \emptyset $} \\
\multicolumn{2}{|p{6.8cm}|}{} \\
\multicolumn{2}{|p{6.8cm}|}{\emph{Generation}} \\
\multicolumn{1}{|p{2cm}|}{$\textsc{Replace}$} & \multicolumn{1}{p{4.8cm}|}{$B = $ \{ Brazilian players entered the penalty} \\
\multicolumn{1}{|p{2cm}|}{$(B, \text{his}, \text{Ruiz})$} & \multicolumn{1}{p{4.8cm}|}{area before \textcolor{red}{Ruiz 's} kick .\}} \\

\hline \multicolumn{2}{p{6.8cm}}{\centering{\normalsize{\rotatebox[origin=c]{-90}{\MVRightarrow}}}} \\
\hline 
\multicolumn{2}{|p{6.8cm}|}{\textbf{4. Output sentence pair}} \\
\multicolumn{2}{|p{6.8cm}|}{$A = $ \{\textcolor{red}{Ruiz} ordered his first shot to be retaken . \}} \\
\multicolumn{2}{|p{6.8cm}|}{$B = $ \{Brazilian players entered the penalty area before \textcolor{red}{Ruiz 's} kick . \}} \\
\hline
\end{tabular}}
\end{center}
\caption{Detailed two-rule execution example. We show in \textcolor{red}{red} parts of the input that are used for detection or modified during execution. The input token list $Z$ is of a single sentence. First, the rule for inner connective is applied, splitting $Z$ into two sentences $A,B$, without the connective \nl{because}. Then, applying the anaphora rule, the pronoun \nl{his} in $B$ is replaced  with the entity it refers to in $A$, to obtain a new sentence pair.}
\label{tab:rule_example}
\end{table}

\begin{figure}[h!]
    \centering
    \includegraphics[scale=0.35]{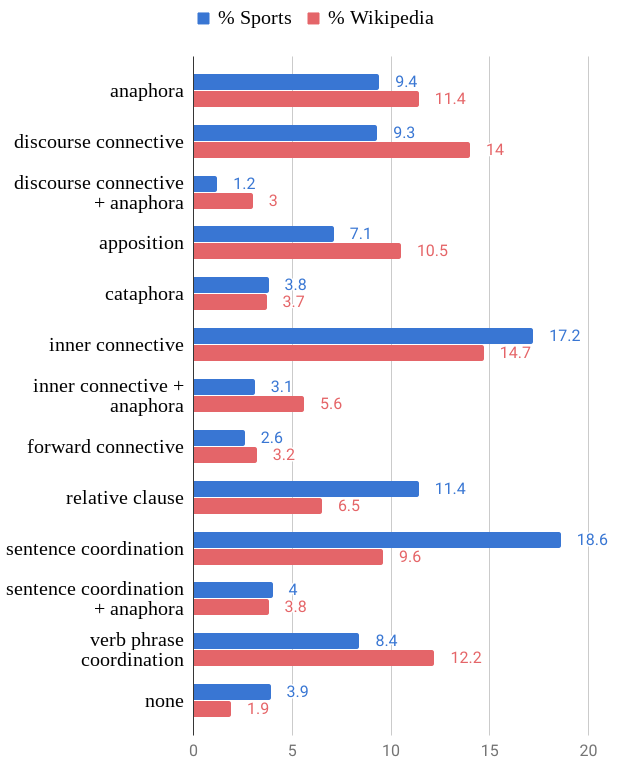}
    \caption{Discourse type distribution of the sports and Wikipedia portions of \discofuse{} after down-sampling.}
    \label{fig:discourse_type_distribution}
\end{figure}

\begin{table}[h!]
\begin{center}
\footnotesize{
\begin{tabular}{|p{1.6cm}|p{0.6cm}||p{1.6cm}|p{0.6cm}|}
\multicolumn{2}{c}{\textsc{Sports}} & \multicolumn{2}{c}{\textsc{Wikipedia}} \\
\hline
 & \% &  & \% \\
\hline
and & 12.0 & and & 12.5 \\
\hline
but & 10.9 & but & 10.7 \\
\hline
because & 8.1 & although & 8.4 \\
\hline
although & 5.4 & however & 8.2 \\
\hline
so & 4.9 & because & 7.7 \\
\hline
or & 4.8 & so that & 2.1 \\
\hline
however & 3.5 & while & 2.0 \\
\hline
while & 2.4 & or & 1.8 \\
\hline
so that & 2.2 & so & 1.2 \\
\hline
unless & 2.2 & for example & 1.0 \\
\toprule
\end{tabular}}
\end{center}
\caption{Most common connectives in 
  \discofuse{} after down-sampling. Percentages are
  with respect to the entire dataset, including examples without a connective.}
\label{tab:connectives}
\end{table}

\subsection{\discofuse{} Data Distribution}
\label{subsec:data_distribution}

Figure~\ref{fig:discourse_type_distribution} and Table~\ref{tab:connectives} show the distributions of discourse types and most common connectives in the two parts of \discofuse{}. 

Analyzing the dataset reveals significant differences
in discourse phenomena between the two types of documents (Figure~\ref{fig:discourse_type_distribution}).
E.g., coordination is very common in Wikipedia while anaphora is dominant in Sports. Likewise, the distribution of discourse connectives is quite different (Table~\ref{tab:connectives}).


\subsection{Neural Models Parameters}
\label{subsec:supplemental_neural}
The models \dfs{}, \dfw{}, \dfsw{} share the same Transformer network architecture, that was originally proposed by \citet{vaswani2017attention}. During training, we split the samples to buckets by their text length, and use different batch size between 60-100 for each bucket. Further configuration details and the hyperparameters used for training of each model are provided in Table~\ref{tab:model_parameters}. 

\begin{table}[ht]
\begin{center}
\scriptsize{
\begin{tabular}{lccc}
\multicolumn{1}{c}{} & \textbf{\dfs{}} & \textbf{\dfw{}} & \multicolumn{1}{c}{\textbf{\dfsw{}}} \\
\hline
number of hidden layers & 7 & 7 & 7 \\
hidden dimension & 1024 & 1024 & 1024 \\
filter size & 2048 & 2048 & 2048 \\
number of heads & 16 & 16 & 16 \\
beam width & 4 & 4 & 4 \\
attention dropout rate & 0.1 & 0.2 & 0.2 \\
ReLU dropout rate & 0.4 & 0.3 & 0.2 \\
learning rate & 0.14 & 0.07 & 0.11 \\
\hline
\end{tabular}}
\end{center}
\caption{Parameters and hyperparameters of the models \dfs{}, \dfw{}, \dfsw{}.}
\label{tab:model_parameters}
\end{table}

\begin{table}[ht]
\begin{center}
\footnotesize{
\begin{tabular}{lcc}
\multicolumn{1}{c}{} &  \\
\hline
number of encoder layers & 3 \\
number of decoder layers & 1 \\
hidden dimension & 128 \\
beam width & 20 \\
scheduled sampling probability & 0.2 \\
dropout rate & 0.2 \\
learning rate & 0.001 \\
learning rate decay & 0.98 \\
\hline
\end{tabular}}
\end{center}
\caption{Parameters and hyperparameters of the CopyNet models used for transfer learning.}
\label{tab:copynet_model_parameters}
\end{table}

In our transfer learning experiment, we trained two CopyNet models \cite{gu2016copying}: a model trained on \websplit alone, and a model pretrained on \dfw{} and finetuned on \websplit.
The first model was trained for 200,000 steps on \websplit{}, whereas the second model was pretrained for 1 million steps on \dfw{} and then finetuned for 100,000 steps on \websplit. Again, the samples were split to buckets by their text length, with batch sizes between 25-125 for each bucket. 
The final test scores were computed with the parameters that maximize the validation \sari{} score during training.
The network architecture and hyperparameters were shared between the models and not optimized during training. 
They are listed in Table~\ref{tab:copynet_model_parameters}.


\end{document}